\documentclass[11pt,a4paper]{article}
\usepackage[hyperref]{acl2021}
\usepackage{times}
\usepackage{latexsym}

\usepackage{dirtytalk}
\usepackage{microtype}
\usepackage{graphicx}
\usepackage{array}
\newcolumntype{H}{>{\setbox0=\hbox\bgroup}c<{\egroup}@{}}

\usepackage{booktabs}
\usepackage{multirow}
\usepackage{dirtytalk}

\aclfinalcopy 

\title{Lost in Space Marking}

\author{Cassandra L. Jacobs \\
  Department of Linguistics \\
  University at Buffalo \\
  Buffalo, NY, USA \\
  \texttt{cxjacobs@buffalo.edu} \\\And
  Yuval Pinter \\
  Department of Computer Science \\
  Ben-Gurion University of the Negev \\
  Beer Sheva, Israel \\
  \texttt{uvp@cs.bgu.ac.il} \\}

\date{}

\begin{document}
\maketitle
\begin{abstract}
We look at a decision taken early in training a subword tokenizer, namely whether it should be the word-initial token that carries a special mark, or the word-final one.
Based on surface-level considerations of efficiency and cohesion, as well as morphological coverage, we find that a Unigram LM tokenizer trained on pre-tokenized English text is better off marking the word-initial token, while one trained on raw text benefits from marking word ends.
Our findings generalize across domains.
\end{abstract}

\section{Introduction}
\label{sec:intro}

Modern NLP is dominated by large pre-trained models, systems which are large, complex, and costly to train.
As a result, much research effort is put into questions of tuning and configuring the various layers and training regimes for improving prediction quality on a growing number of tasks~\cite{rogers2020primer}.
Unfortunately, not as much research asks questions about the decisions made at the most upstream parts of the models, those that deal with input tokenization and subword vocabulary creation.

In this exploratory work, we isolate a single decision point which appears to be resolved arbitrarily by existing model developers, with no consensus but also no underlying theory:
\textbf{should subword tokenizers mark word boundaries at the beginning or the end?}
The immediate effects of such a decision are easy to describe: an out-of-vocabulary word\footnote{We use the term \textit{word} to refer to a space-delimited string of characters, post-pre-tokenization if such a process exists. Consequentially, \textit{subword} refers to tokens which are either word-length or form a proper substring of a word.} which is represented by multiple tokens, or \say{pieces}, e.g.~the plural noun \textit{polynomials}, shares the token \textit{polynomial} with its singular form in initial-boundary-marking tokenizers, but not in final-boundary-marking ones.
Conversely, a word prepended by a prefix like \textit{neoconservative} shares the stem token in final-boundary-marking tokenizers and not in the initial variants.
In lexical compounds, the boundary marking decision can affect whether \textit{thunderstorm} shares a token with its internal head, \textit{storm}, or its modifier \textit{thunder}.
The consequence for the unshared stem, in such cases, is often further segmentation since the suffix \textit{storm} (or prefix \textit{thunder}) rarely appears in the training corpus in a similar boundary-relative position.

It might appear at first that in English, a heavily-suffixing language \cite{wals-26}, a tokenizer that marks word-initial pieces is more appropriate.
Such a claim requires empirical support, but consideration of common practice can also be offered to challenge it:
for one, pre-tokenization such as punctuation separation and accent normalization is not always applied consistently when moving on to a downstream text.
A model that was trained on untreated text may find it difficult to process an NER dataset (for example) where punctuation is separated from preceding words, rendering a word-final-marking tokenizer more robust to change;
some tokenizers like BERT's Wordpiece~\cite{devlin-etal-2019-bert} \say{mark} a class of tokens by omission, i.e.~marking the non-initial pieces rather than initial ones.
This discrepancy surfaces edge case effects when compared with a seemingly-equivalent tokenizer like GPT-2's~\cite{radford2019language}, which marks initial pieces but only if they are prepended by a space (i.e.~not sentence-initial pieces).
We survey models' decisions for this question in \autoref{tab:systems}.

\begin{table}
    \centering
    \small
    \begin{tabular}{ccc} \toprule
         & Specify word-initial & Specify word-final \\
         \midrule
        Boundary & ALBERT, RoBERTa, & GPT, XLM \\
        mark & XLNet, BART$^*$, T5, & \\
        & MarianNMT, GPT-2$^*$ & \\
        Internal & \multirow{2}{*}{BERT, mBERT} & \multirow{2}{*}{Subword-NMT$^\dagger$} \\
        mark & &  \\
         \bottomrule
    \end{tabular}
    \caption{Models of various marking decisions. Implementations are taken from the Huggingface Tokenizers package~\cite{wolf-etal-2020-transformers}, except for $^\dagger$\cite{sennrich-etal-2016-neural}.
    $^*$These tokenizers default to not beginning sentences with spaces, making sentence-initial pieces equivalently marked as word-non-initial pieces.}
    \label{tab:systems}
\end{table}

We offer three courses of analysis for the English Unigram LM tokenizer~\cite{kudo-2018-subword}, recently found to outperform BPE~\cite{sennrich-etal-2016-neural} on both statistical measures and downstream performance~\cite{bostrom-durrett-2020-byte}.
We conduct type-level, token-level and corpus-level statistical analysis on the initial-vs.-final variable~(\S\ref{sec:prop},\ref{sec:stat}), and add a morphological coverage test (\S\ref{sec:morph}) motivated by the question of subword models' ability to pick up meaningful linguistic generalizations using corpus statistics only.
We find that: (a) pre-tokenizing input text affects the boundary marking decision: unprocessed corpora produce better final-marking vocabularies, whereas pretokenized ones produce better initial-marking vocabularies;
(b) corpus compression efficiency trades off against information-theoretic measures;
(c) more morphemes are detected by final-marking vocabularies, but the techniques are complementary to each other and finding a combination of both can prove useful.


\section{Model Properties}
\label{sec:prop}

The unigram language model~\cite[Unigram LM,][]{kudo-2018-subword} is a method for creating subword vocabularies by optimizing a frequency objective over a large corpus:
a vocabulary 
is initialized to include all subword sequences in the corpus above a frequency threshold, and then portions of the vocabulary are removed iteratively such that the likelihood scores for corpus-level word segmentations are minimally affected, until a pre-defined vocabulary size has been reached.
In the final vocabulary, each piece is assigned a score which is simply the log-probability of its appearance in the training corpus given the vocabulary, allowing for a decoding algorithm to select the most probable segmentation for a space-delimited word.

We train four Unigram LM models on a pre-specified corpus of 1 million sentences randomly sampled from the March 2019 dump of English Wikipedia.
We use the Sentencepiece package\footnote{Version 0.1.91, \url{https://github.com/google/sentencepiece}} with full character coverage, trimming the piece vocabulary by a factor of 75\% iteratively until reaching a size of 32,000.
Our evaluated conditions are whether or not the text is pretokenized (using version 3.7 of the StanfordNLP tokenizer~\cite{qi2018universal}; no other normalization, such as lowercasing, is performed), and whether the word-initial or word-final piece is marked:\footnote{We accomplish this by simply reversing all training and inference texts, and applying the model as-is.}
\textsc{RawInit}: no pretokenization, word-initial marked;
\textsc{RawFin}: no pretokenization, word-final marked;
\textsc{StanInit}: Stanford pretokenization, word-initial marked;
\textsc{StanFin}: Stanford pretokenization, word-final marked.

At the type level, we note that normalized unigram frequencies are encoded as piece scores in the model artifact, so we can compute the entropy of the piece distributions.
A better-balanced model should present higher entropy, although this may also be the artifact of excessive splitting of frequent character sequences into several \say{cases}.
We find that under both pretokenization conditions, the \textsc{Fin} condition produces higher-entropy distributions: 7.549 vs. 7.509 on \textsc{Raw}, 7.400 vs. 7.285 on \textsc{Stan}.

\begin{figure*}
    \centering
    \begin{tabular}{cc}
        \includegraphics[width=7.5cm,height=4cm]{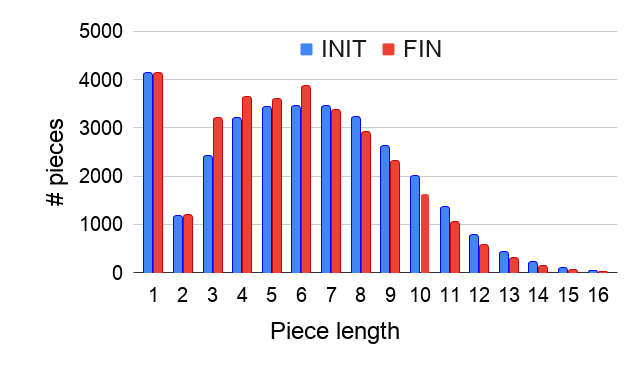}
         & 
         \includegraphics[width=7.5cm,height=4cm]{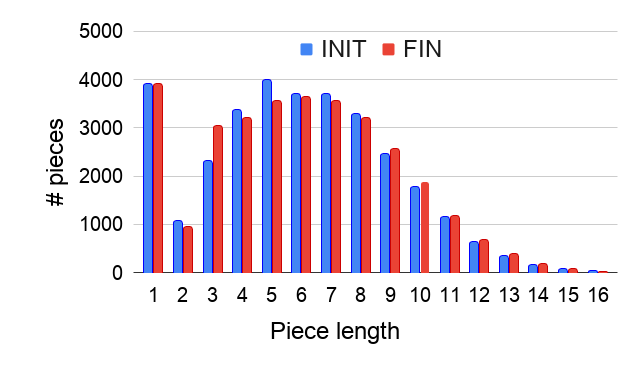}     
         \\
    \end{tabular}
    \caption{Token length distributions for models trained on (left) raw, (right) Stanford-pretokenized samples.}
    \label{fig:lengths}
\end{figure*}

Next, we consider the distributions of piece character lengths.
Intuitively, a model which finds more long pieces can more efficiently compress the information encoded in the corpus; \citet{bostrom-durrett-2020-byte} argued that a distribution with mean centered around 6 resembles that of English morphemes better.
As demonstrated in \autoref{fig:lengths}, these two arguments are at odds in our experiment: in the \textsc{Raw} condition, the final-marking model produces shorter pieces, but centered at 6.
In the \textsc{Stan} condition, the situation is reversed.

\section{Corpus analysis}
\label{sec:stat}

Our next evaluation considers the utility of each marking schema in the context of a large corpus.
We report several surface-level statistics which are not immediately translatable to downstream model efficacy, but are \say{clean} in the sense that they introduce no stochastic noise other than corpus selection, and may still indicate a desired preference for downstream model inputs.
We measure each statistic on the same \textbf{training} corpus as initially passed to the tokenizer; on a different sample of \textbf{in-domain text} from the same Wikipedia dump; and on an \textbf{out of domain} sample from Reddit's \texttt{r/politics} feed extracted in the first half of 2015.

\paragraph{Token ratio.}
We count the total pieces tokenized by the model, as a measure for efficient text encoding, and divide by the number of space-delimited words in the corpus (lower = better).

\paragraph{Corpus-level entropy.}
We distill the piece-type distribution across each corpus into a single entropy number, under the assumption that a good encoding balances token distribution out rather than relying on a \say{heavy head} of certain types (higher = better).

\paragraph{Bigram LM perplexity.}
We train a basic bigram language model with add-$\epsilon$ smoothing on each tokenizer's original training corpus and apply it to each inference corpus, measuring overall base-2 perplexity.
We expect differences between \textsc{Init} and \textsc{Fin} to be mainly the result of in-word prediction ability.
We use $\epsilon\leftarrow 0.005$ (lower = better).


\begin{table*}
    \centering
    \small
    \begin{tabular}{lHcHcccHcccHcc} \toprule
        Model & Training & \multicolumn{4}{c}{Training corpus} & \multicolumn{4}{c}{In-domain corpus} & \multicolumn{4}{c}{Out-of-domain corpus} \\
        & & Toks/word & change & Entropy & Ppl. & Toks/word & change & Entropy & Ppl. & Toks/word & change & Entropy & Ppl. \\
        \midrule
        \textsc{RawInit} & N/A & \textbf{1.356} & & \textbf{7.591} & 174.89 & \textbf{1.358} & & \textbf{7.591} & 276.57 & \textbf{1.440} & & \textbf{6.910} & 586.41 \\
        \textsc{RawFin} & N/A & 1.393 & 2.67\% & 7.648 & \textbf{165.54} & 1.394 & 2.66\% & 7.648 & \textbf{253.28} & 1.457 & 1.17\% & 7.052 & \textbf{563.25} \\
        \midrule
        \textsc{StanInit} & N/A & 1.208 & & \textbf{7.361} & \textbf{153.41} & 1.409 & & \textbf{7.358} & \textbf{262.98} & 1.490 & & \textbf{6.773} & \textbf{535.80} \\
        \textsc{StanFin} & N/A & \textbf{1.200} & -0.64\% & 7.471 & 157.76 & \textbf{1.400} & -0.64\% & 7.468 & 274.38 & \textbf{1.471} & -1.25\% & 6.883 & 579.09 \\
        \bottomrule
    \end{tabular}
    \caption{Corpus-level measures of different Unigram LM vocabularies on different text sources.}
    \label{tab:corpus_stats}
\end{table*}

Our results, presented in \autoref{tab:corpus_stats}, show a number trends, all of which remain consistent across corpora and domains.
While type-level entropy is always lower in the \textsc{Init} marking scheme, other findings suggest a trade-offs between compression ability (better for \textsc{Init} on raw text but for \textsc{Fin} when pre-tokenized) and prediction regularity (vice versa).
It may be the case that final-marked tokens are easier to predict after non-final tokens in a left-to-right bigram language model, suggesting these results may be particularly important for downstream regressive models like GPT-$n$, as opposed to masked language models.

\section{Morphological recovery}
\label{sec:morph}


Given the difference in segmentation strategy, and ultimately in the difference in the quality of the representations, 
we might ask how well both the initial- and final-marking implementations of Unigram LM recover morphologically well-formed subword sequences. 
To that end, we consulted the morphological section of the English Lexicon Project \newcite{balota2007english} available from the \texttt{citylex} repository.\footnote{\url{https://github.com/kylebgorman/citylex}} 
We extracted morphological information from all words in the database, identifying bound (e.g., \say{-ing}) and unbound morphemes (e.g., \say{cat}), resulting in a \say{gold} set of 18,794 strings corresponding to linguistically-motivated morphological forms.

\begin{figure*}
    \centering
    \begin{tabular}{cc}
        \includegraphics[width=7.5cm,height=4cm]{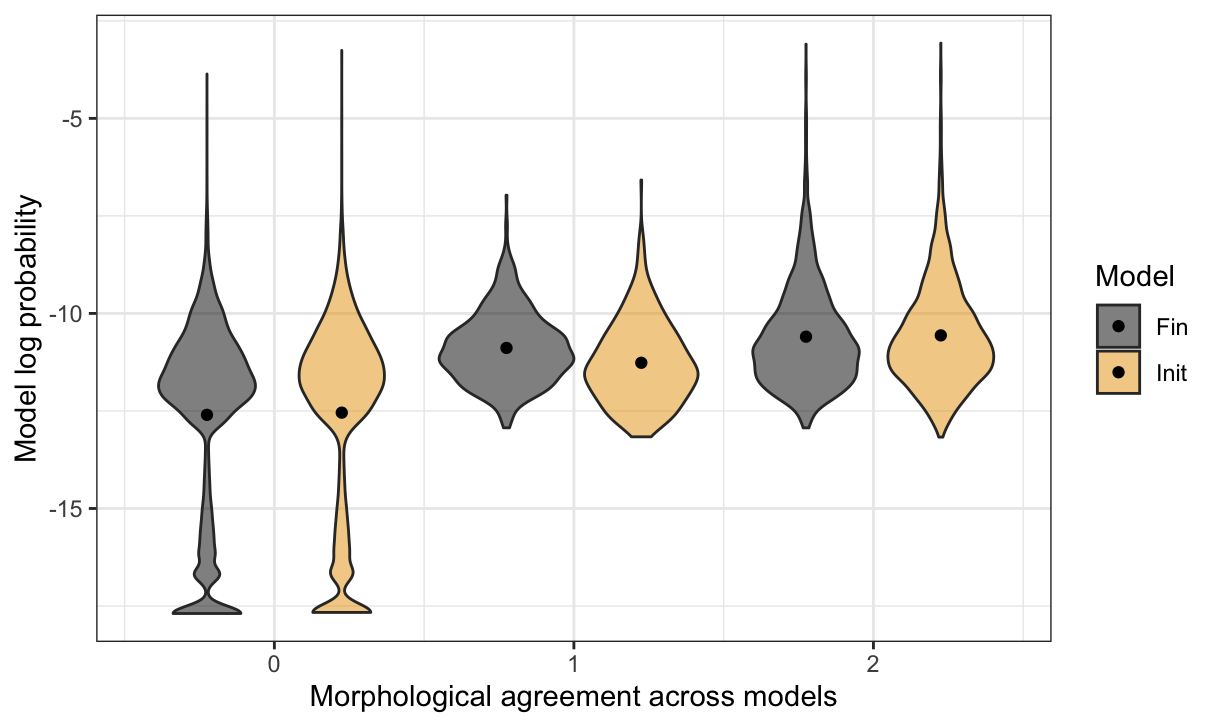}
         & 
        \includegraphics[width=7.5cm,height=4cm]{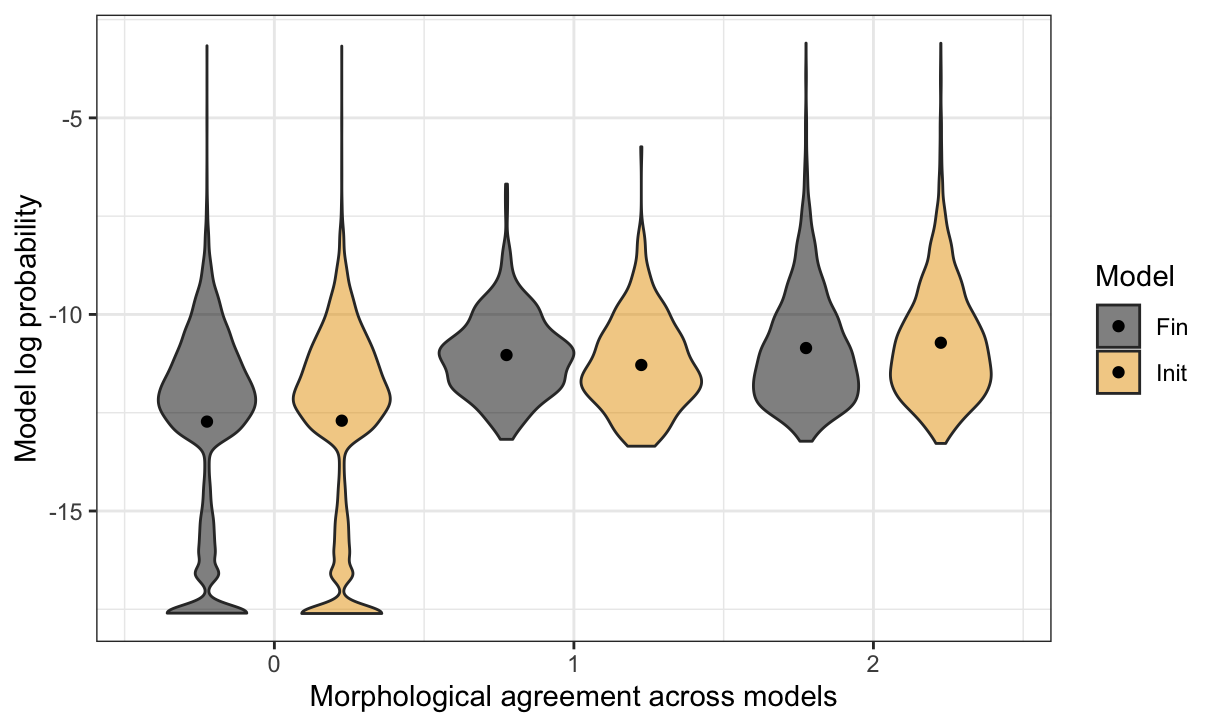}     
         \\
    \end{tabular}
    \caption{Unigram LM log-probabilities for (left) raw, (right) Stanford-pretokenized subwords, by morphology.}
    \label{fig:scores}
\end{figure*}



\paragraph{Coverage.} 
Tokenization from sentences has a sizeable effect on the performance of subword vocabularies on perplexity. 
It is feasible that pre-tokenization will likewise influence morphological recovery. 
The \textsc{RawInit} model recovered 5,170 (27\%) morphemes, while \textsc{RawFin} models recovered 6,049 (32\%), suggesting that initial-marking can be detrimental to morphological recovery. 
Assessing the union of the subword vocabularies learned by both models reveals 7,898 (42\%) morphological subwords. 
Some morphemes that were identified only by \textsc{RawInit} included 109 rarer initial morphemes (e.g., \textit{co-}), while \textsc{RawFin} identified some 482 non-initial morphemes that likely came from compounds (e.g., \textit{-office}, \textit{-league}).
We find similar but less stark findings for the pre-tokenized text, where the \textsc{StanInit} model recovered 5,391 morphemes (28\%), against 5,537 (29\%) for \textsc{StanFin}, with a union of 7,568 covered morphemes (40\%).

\paragraph{Likelihood.}


We compared the log probabilities that the \textsc{Init} and \textsc{Fin} models assign to each subword.
We sought to test whether higher language model scores are allocated to subword strings that are probable morphemes than to non-morphemes.
First, we characterized morphological agreement between the models as a single number: 0 (neither model identifies a subword as a morpheme); 1 (recognized as a morpheme in exactly one model); and 2 (both models identify the morpheme) as possible values.
We used a linear regression to predict subword tokens' log probabilities as a function of (a) morphological agreement score (0-2); and (b) the identity of the tokenization model from which the log probability was taken.
The model presented significantly higher log probabilities at higher agreement levels: a mean value of -10.7 for the 2 class, vs. -11.3 and -12.7 for (1) and (0) respectively ($\beta$ = 0.95, $t$ = 43.5, $p$ $<$ .001), but this was true to similar degrees for both the \textsc{Init} or \textsc{Fin} model.
The results are qualitatively similar for both the pretokenized and raw corpora.
We visualize the data used in these analyses in \autoref{fig:scores}, which shows the probabilities assigned by each model depending on a subword's morphological status.

Altogether, the results of these experiments suggest that the greatest power is in assessing the overlap between the \textsc{Init} and \textsc{Fin} models --- significant morphological faithfulness can be achieved by integrating both vocabularies.

\section{Related Work}
Little prior work has compared different tokenization schemes.
\citet{bostrom-durrett-2020-byte} compared the standard BPE implementation in BERT to a subword vocabulary learned using the Unigram LM algorithm, finding that Unigram LM was better able to identify morphological segments.
In a study comparing BPE to Morfessor, \citet{banerjee2018meaningless} found that combining the two segmentation algorithms produced a superior subword vocabulary for machine translation.
In a study of the usefulness of subword segmentation in Hebrew, \newcite{klein2020getting} found that forcing a morphologically-informed parsing scheme, rather than relying on word pieces from multilingual BERT \cite{devlin-etal-2019-bert}, improved performance on tagging accuracy.
However, while the morphological typology of a language can play a significant role in the success of a language model, recent work suggests that morphological variability is less important than other types of variability \cite{mielke-etal-2019-kind}.

\section{Conclusion}

We find that the oft-ignored decision of how to mark subword tokens for the sake of word recoverability carries noticeable implications to the efficacy of the resulting model's representation of vocabularies and corpora, as well as the subwords' correspondence to morphemes.
This variable interacts in a nontrivial manner with the degree to which the training corpus has been pre-processed.
We urge researchers to pay heed to these preliminary decisions, and make their reasoning or supporting experimental evidence explicit when reporting model performance.
One promising avenue for future work may be to try and synthesize initial- and final-word marking into one tokenization model which can recover a large amount of morphemes, while still benefiting from the compression abilities of both techniques.


\bibliographystyle{acl_natbib}
\bibliography{anthology,mark_bpe_end}


\end{document}